\documentclass{article}
\usepackage[utf8]{inputenc}
\usepackage{graphicx}
\usepackage{subcaption}
\usepackage{float}
\usepackage[table]{xcolor}
\usepackage[super]{nth}
\usepackage[backend=biber, sorting=none]{biblatex}
\usepackage[margin=1in]{geometry}
\usepackage{authblk}
\usepackage{url}

\title{\textbf{Temporal Analysis on Topics Using Word2Vec}}
\date{}

\makeatletter
\newcommand\email[2][]%
   {\newaffiltrue\let\AB@blk@and\AB@pand
      \if\relax#1\relax\def\AB@note{\AB@thenote}\else\def\AB@note{\relax}%
        \setcounter{Maxaffil}{0}\fi
      
      \begingroup
        \let\protect\@unexpandable@protect
        
        \def\thanks{\protect\thanks}\def\footnote{\protect\footnote}%
        \@temptokena=\expandafter{\AB@authors}%
        {\def\\{\protect\\\protect\Affilfont}\xdef\AB@temp{#2}}%
         \xdef\AB@authors{\the\@temptokena\AB@las\AB@au@str
         \protect\\[\affilsep]\protect\Affilfont\AB@temp}%
         \gdef\AB@las{}\gdef\AB@au@str{}%
        {\def\\{, \ignorespaces}\xdef\AB@temp{#2}}%
        \@temptokena=\expandafter{\AB@affillist}%
        \xdef\AB@affillist{\the\@temptokena \AB@affilsep
          \AB@affilnote{}\protect\Affilfont\AB@temp}%
      \endgroup
      
       \let\AB@affilsep\AB@affilsepx
}
\makeatother

\author[1]{\textbf{Angad Sandhu}}
\author[2]{\textbf{Aneesh Edara}}
\author[2]{\textbf{Vishesh Narayan}}
\author[2]{\textbf{Faizan Wajid}}
\author[2]{\textbf{Ashok Agrawala}}
\affil[1]{Department of Computer Science, Johns Hopkins University}
\affil[2]{Department of Computer Science, University of Maryland}

\begin{document}
\maketitle

\begin{abstract}
The present study proposes a novel method of trend detection and visualization — more specifically, modeling the change in a topic over time. Where current models used for the identification and visualization of trends only convey the popularity of a singular word based on stochastic counting of usage, the approach in the present study illustrates the popularity and direction that a topic is moving in. The direction in this case is a distinct subtopic within the selected corpus. Such trends are generated by modeling the movement of a topic by using \textit{k}-means clustering and cosine similarity to group the distances between clusters over time. In a convergent scenario, it can be inferred that the topics as a whole are meshing (tokens between topics, becoming interchangeable). On the contrary, a divergent scenario would imply that each topics' respective tokens would not be found in the same context (the words are increasingly different to each other). The methodology was tested on a group of articles from various media houses present in the 20 Newsgroups dataset. 

\end{abstract}


\section{Introduction}
Researchers interested in determining the features and dynamics of a specific environment have access to vast quantities of digital data from which to derive answers to their inquiries. The amount of written news content available on the internet alone provides a plethora of knowledge, context, and documentation that may be utilized to solve social-scientific issues. Even at the computational level, sifting through this amount of data and highlighting conversations and patterns of special interest is difficult, let alone for individual scholars to explore. Topic models are Bayesian statistical models that have proven their accuracy in many applications \cite{textmining}. Given a large corpus, these models permit the extraction of topics that structure the texts, and the topics themselves can be simplified to a list of keywords. Dirichlet-multinomial regression (DMR) topic model \cite{mimno2012topic} with document features such as author, references, dates, etc. can enhance the performance of such topic models.

\section{Related Works}

The present study aims to measure how the robustness of terms, derived from observed topics occurring in the corpus manually, may evolve temporally and correspond with known events described in the corpus' component papers. 

In comparison to a study of current relevant methodologies, this analysis borrows and improves on previous methods to provide three key strengths in a novel approach to solving the stated problem: Capturing meaningful language use beyond the level of a word, the use of predefined topics to empower framework-based analysis and considering the temporal change in n-gram usage across a corpus of documents over a time period of interest. Kherwa and Bansal \cite{Kherwangram}, though using a topic modeling approach, similarly harness the semantic power of n-grams; they cite the difference between the words ``New'' and ``York'' occurring separately and ``New York'' as a bigram. A term-frequency measurement similar to Don et al. \cite{59394777347e41faa5b6431040808dfb} is applied, but scaled to the level of a corpus, as opposed to single documents with distinct models. Ahmed, Traore, and Saad \cite{Ahmed2017DetectionOO} note an epistemological limitation in their analysis: Identification of terms and trends of interest in text data typically requires some prior knowledge of what is to be found.

Bamler et al. \cite{bamler2017dynamic} presented a language model for probabilistic temporal text data which tracks the semantic evolution of individual words over time. The model uses Word2Vec \cite{mikolov2013efficient} to create embeddings, which are connected through inference algorithms that allow training the model jointly over all time periods. Word embedding trends are more interpretable and lead to higher predictive likelihoods than competing methods that are based on static models trained separately on time slices. This method is used to evaluate the change in the semantics of each word over time.

Topic modeling is useful for discovering latent topics that can later be associated with known trends or events. To attribute text in a document to a specific topic, LDA \cite{10.5555/944919.944937} is utilised. It creates a topic per document model and a words per topic model using Dirichlet distributions as the modelling framework. However, The approach in the present study employs a novel analysis, i.e hypothesizing that given a known set of topics, emergent trends or events can be discovered by tracking the temporal change between terms. Insights synthesized from the aforementioned research and development of a novel approach are used to test this hypothesis.

\section{Background}
When deciding what algorithms to use to model trends, the first consideration was towards the final output that needs to reach the end-user so they can gain insights. In the present study, the idea was to create a framework that would not only detect how much a topic evolves, but also in which way it is evolving semantically. In addition to the idea of giving the users flexibility in choosing their topic that is the ``\textit{umbrella}'' term, the users can decide the \textit{sub-topics} within the umbrella term as well. Thus, word embeddings and \textit{k}-means clustering became the immediate choices as the semantic changes in a word could be modeled over time and also for generating topics.

\subsection{Word2Vec Context}
Word2Vec \cite{mikolov2013distributed} can utilize either of two model architectures to produce a distributed representation of words: continuous \textit{bag-of-words} (CBoW) or continuous \textit{skip-gram}. In the continuous bag-of-words architecture, the model predicts the current word from a window of surrounding context words (window size). The order of context words does not influence prediction (bag-of-words assumption). In the continuous skip-gram architecture, the model uses the current word to predict the surrounding window of context words. The skip-gram architecture weighs nearby context words more heavily than most distant context words. The need for word prediction when the goal is to generate embeddings comes from the nature of language which is unlabeled. To overcome this, an artificial task is created for the neural network: predict a targeted word from its context and vice versa in a skip-gram. Although the inputs and outputs of the single layer neural network are not the objective, the weights that feed into the final softmax layer will be the eventual embeddings that convey the input word's semantic meaning. The softmax, multinomial distribution can be defined as follows:

\[P( w_j | w_i ) = \frac{exp({v'_{w_j}}^{T}*v_{w_i})}{\sum_{j'=1}^{V} exp({v'_{w_j'}}^{T}*v_{w_i})} \]

Where the probability of the target word w$_{j}$ given the surrounding word(s) w$_{i}$ is maximized. Also note that v$_{w}$ and v'$_{w}$ are two representations of the word ``w''. v$_{w}$ comes from rows of the input hidden weight matrix, and v'$_{w}$ comes from the columns of the hidden output matrix. Using gradient descent, the embeddings can be optimized without having a labeled dataset \cite{word2vecparameterrong}. Since this model is trained in an online setting, the goal is to take a small step mediated by the ``learning rate'' to minimize the distance between the current vectors for w$_{j}$ and w$_{i}$, thereby increasing the probability P(w$_{j}$ $|$ w$_{i}$). By repeating this process over the entire corpus, the vectors for words that habitually co-occur tend to be nudged closer and closer together. By gradually lowering the learning rate, this process converges towards some final state of the vectors. By the Distributional Hypothesis, words with similar distributional properties tend to share aspects of semantic meaning \cite{Firth1957}. For example, sentences in the corpus such as ``I like to play X with my friends,'' where X is the target word 'w$_{j}$' may be names of sports or activities that are semantically related.

\subsection{\textit{k}-means Clustering Context}
\textit{k}-means clustering is used to give the end-user the flexibility of choosing the subset of information that they may want to target. The user picks a single word topic, which then is converted to a vector, which in-turn is used to find the 100 words closest to it angle-wise with cosine similarity. \textit{k}-means clustering makes this simple because it essentially finds the center of mass (centroid) of a topic without having the rigid constraints of a Latent Dirichlet Allocation (LDA) \cite{mimno2012topic}. The \textit{k}-means clustering when mathematically depicted below computes the objective/squared error function that minimizes the distance from centroid to all other data points in a given cluster (\(m_n\) to all the \(x_n\) word topics). 

\[J = \sum_{n=1}^{N}\sum_{k=1}^{K} r_{nk} * {||x_n - m_n||}^2\]

\section{Methodology}
The idea behind the present study being that, in an environment where comparable corpora (belonging to the same topic or sub-topic) are generated in a controlled and steady manner—such as a fashion magazine or financial newspaper—the more a word/topic is talked about, the more that term will be closely related to the umbrella field. Such as in a sports magazine; as the Olympics come closer, track and field will be much more closely related to the umbrella term of sports itself. 

To demonstrate the effectiveness of the aforementioned framework, two case studies were performed comparing the relation between various themes of trends related to specific and arbitrary events. To implement the above mentioned ideology, the two main steps that were to be taken into account are the preparation of data by carving out the corpora on the basis of time (the time period of comparison being 'months' in present study), hence creating corresponding Word2Vec models for each slice, as well as formulating word trends over all the created models.

\subsection{Dataset}
The \textit{20 Newsgroups dataset} encompasses about 21,000 newsgroup posts or documents on 20 different topics or subtopics (such as atheism, religion, basketball, space etc). Some of the newsgroups are very closely related to each other (e.g. "PC hardware" vs "mac hardware"), while others are highly unrelated (e.g "for sale" vs "Christianity"), making this dataset the optimal choice for the research in this study. The acquired data is split in two subsets: one for training (or development) and the other one for testing (or for performance evaluation). The split between the train and test set is based upon messages posted before and after a specific date. 

\subsection{Data Prepossessing}

Utilizing the 20 Newsgroups dataset, the corpora was loaded up, put into ascending order by date, then first split into smaller splices of data on the basis of years (2016, 2017, 2018 etc.) and further split into their corresponding months. Each corpus being a collection of articles, gives them a formal and structural quality. Therefore, these data points can have a variety of hidden meaning that might be conveyed through semantic analysis. 

The standard data pre-processing steps were performed e.g., keyword tokenization, stopwords and removal of punctuation characters on the dataset. Lemmatization was also performed to get more consistent results. After the data was made adequately usable it was saved to be used later in the Processing step.  

\begin{figure}
  \begin{subfigure}{0.48\textwidth}
    \captionsetup{justification=centering}
    \includegraphics[width=\linewidth]{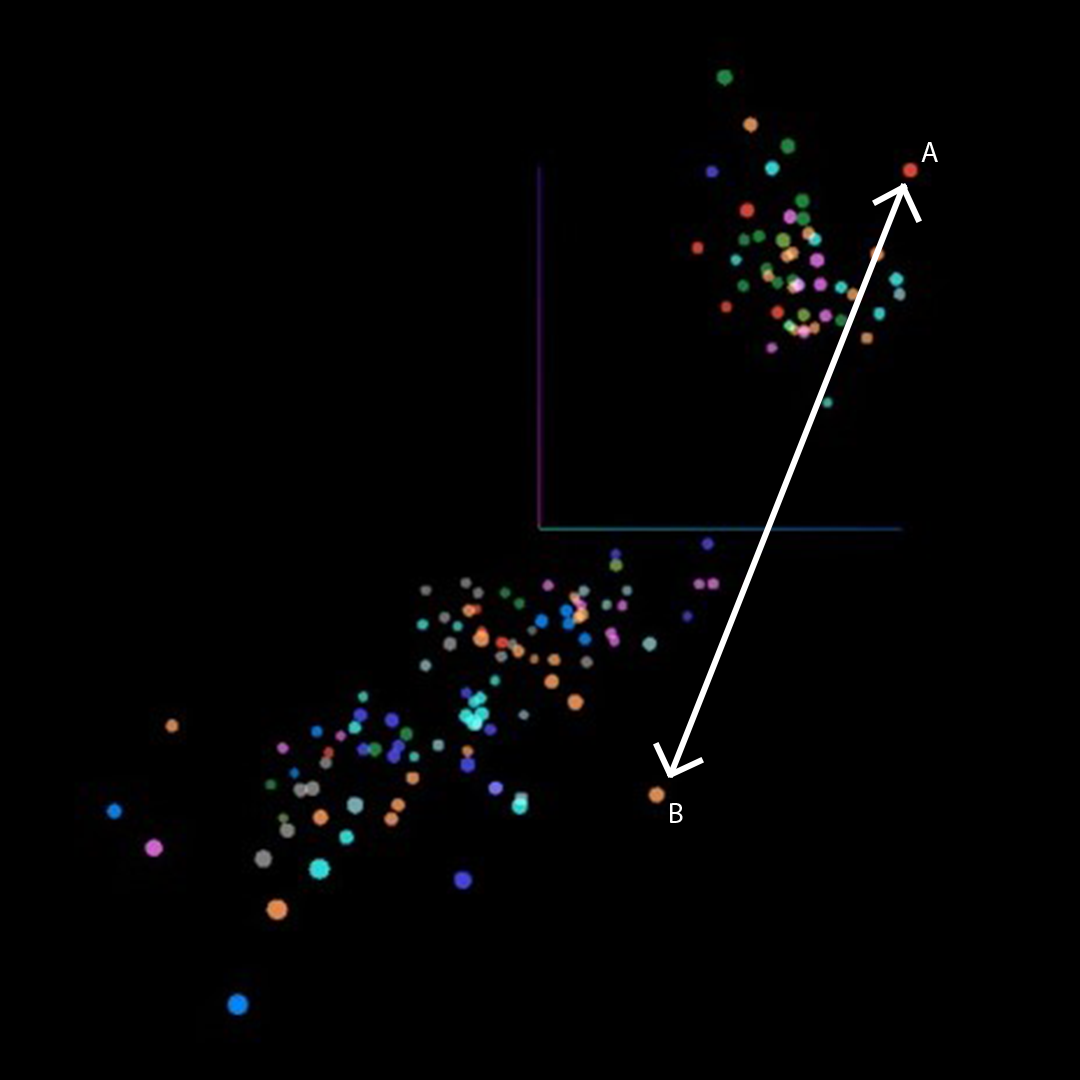}
    \caption{Long distance between the words in Word2Vec model for June 2020} \label{fig:1a}
  \end{subfigure}%
  \hspace*{\fill}   
  \begin{subfigure}{0.48\textwidth}
    \captionsetup{justification=centering}
    \includegraphics[width=\linewidth]{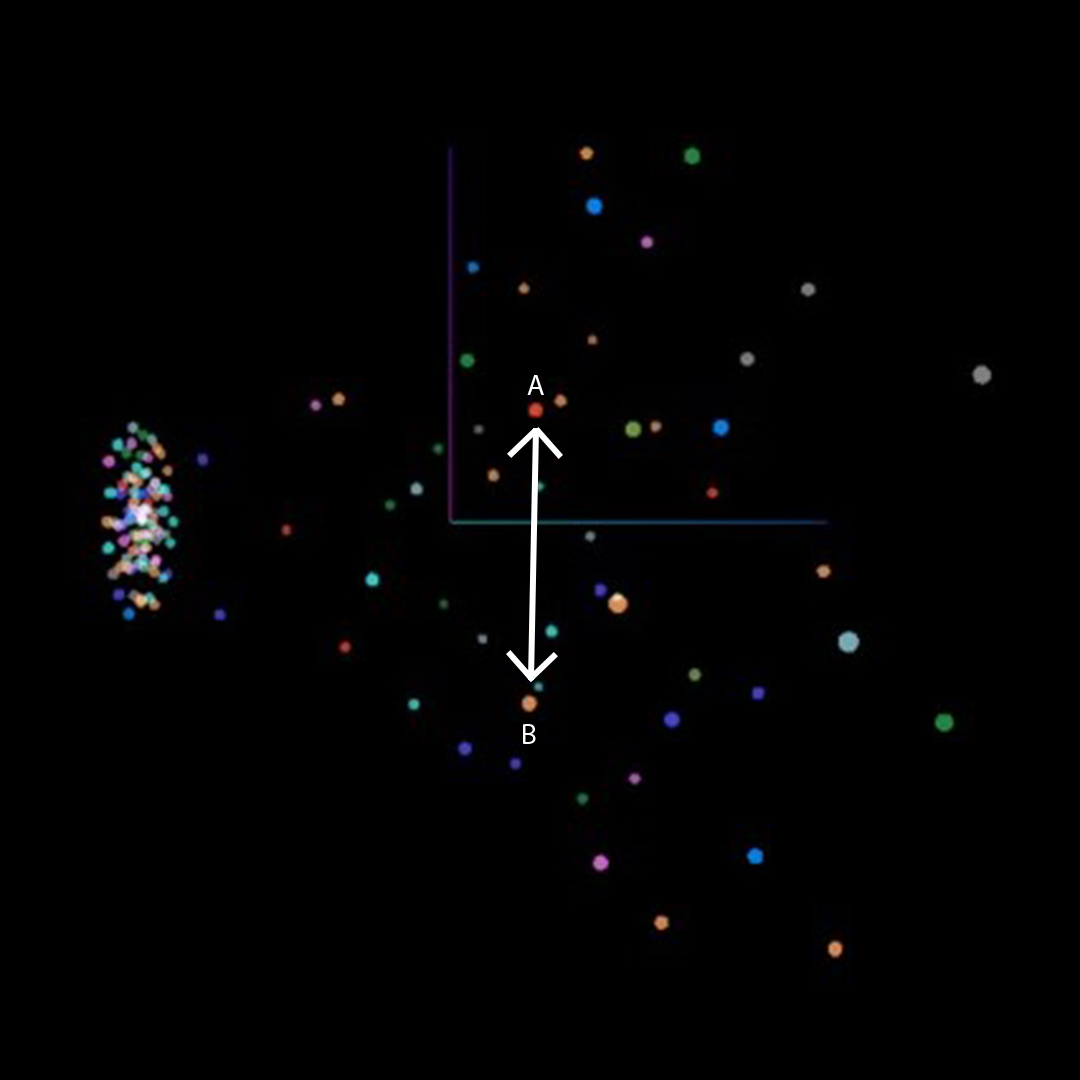}
    \caption{Short distance between the words in Word2Vec model for December 2020} \label{fig:1b}
  \end{subfigure}%

\caption{Comparing semantic similarity between 'CDC' and 'Unofficial' over News Data in 2020} \label{fig:1}
\end{figure}

Before moving onto the next step, \textit{k}-means clustering was done to cross-examine if the data moved as it was hypothesized as. For example, in accordance with expectations, in June 2020, during the first wave of the COVID-19 pandemic, the embeddings of ``Unofficial'' and ``CDC'' (Centers for Disease Control) were at their farthest (Fig: \ref{fig:1a}), but in the following months, they converged. (Fig: \ref{fig:1b})

\subsection{Process}

In the present study, a Continuous Bag of Words Model was trained with a \textbf{worker\_size} of \textbf{5,000}. Training a \textit{GenSim} skip-gram Word2Vec model and tracking the temporal movements of said clusters (refer to Fig. II). As the clusters for each time period are calculated differently, the relative distance between vectors change with accordance to their positioning in clusters over different language models. This very change in distance is an indication of semantic drift, how two separate topics might be more closely related or be less related with each time-step.

The average \textbf{vector\_size} of large text corpora is around 300. In the present study large input corpora was converted into smaller slices, \textbf{vector\_size} is set to 100. We use a window size of 5 to get the best semantic relation between any two terms. \textbf{min\_count} is kept at 10 as well to ignore all words with low usage, such as names of specific officials or names of athletes, only focusing on words with high frequency. After the model for each slice was trained and created, the proceeding step was to measure the trend of words that may or may not be related to the overarching theme of the datasets. 

In the present study, the likeness between a pair of topics is measured by  their cosine similarity, i.e. a number ranging from 0 to 1, where a value close to 0 means that the words are highly correlated (interchangeable) to each other and 1 means that they are never used in the same context. The cosine value is the metric to define the absolute similarity between two words. The maximum cosine value (least correlated) in a pair is highlighted by red whereas the minimum cosine value (most correlated) in a pair is highlighted by green. If a term is not present in the model, its value denoted by '0.0' should be ignored, and is highlighted by blue.

\begin{figure}[H]
    \centering
    \captionsetup{justification=centering}
    \includegraphics[width=1\textwidth]{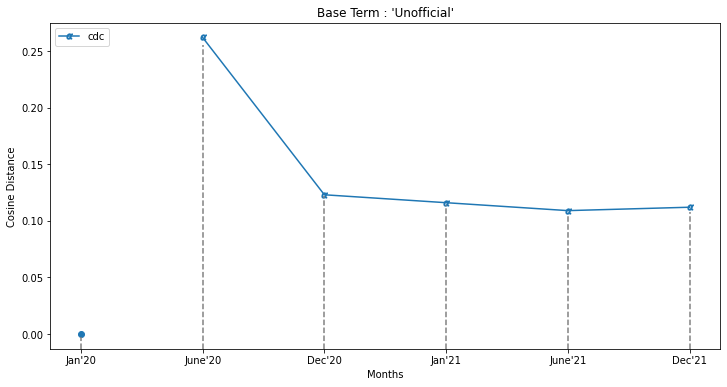}
    \caption{Trend of the terms ``CDC'' through 2020-21 with the relative term ``Unofficial''}
    \label{fig:graph1}
\end{figure}

The first instance of the model (refer to Figure \ref{fig:graph1} and Table \ref{tab:tab1}) shows the trend of a base term with respect to a single relative term. It is the trend of the term ``CDC'' over 2 years, considering the months of January, June and December of 2020 and 2021. The present study considers the base term to be ``unofficial'' to emphasize how the status quo sees the Center of Disease Control. A smaller value of the cosine distance between these two terms suggests that the mainstream considers the organization to be insignificant. In contrast, higher the value, higher will be the people's trust in the organization.

\begin{table}[H]
    \centering
    \captionsetup{justification=centering}
    \begin{tabular}{|c|c|c|c|c|c|c|}
        \hline
        \multicolumn{7}{|c|}{Base Term : \textbf{Unofficial}}\\
        \hline
        Relative Term & \multicolumn{6}{c|}{Timeline (2020 - 2021)}\\
        \hline
        Months &  Jan'20 & Jun'20 & Dec'20 &  Jan'21 & Jun'21 & Dec'21\\
        \hline
        \hline
        CDC &  \cellcolor{blue!25}00.00 & \cellcolor{red!25}0.262 & 0.123 & 0.116 & \cellcolor{green!25}0.109 & 0.112\\
        \hline
    \end{tabular}
    \caption{Cosine Similarity between topics over time on 20 Newsgroups Dataset. Maximum cosine value is red whereas the Minimum cosine value is green.}
    \label{tab:tab1}
\end{table}

As it can be deduced from the data (refer to Figure \ref{fig:graph1} and Table \ref{tab:tab1}), the term ``CDC'' starts being considered from June 2020, corresponding with the sharp increase in COVID-19 cases. A comparatively large cosine distance can be observed between ``CDC'' and ``Unofficial'', conveying that the organization was held in high regard as an entity to look for guidance and information in such a situation. As time progresses to June and December, the public's trust in the ``Media'' and government organizations deteriorate with increasing deaths and growth in the anti-vaccination agenda as well as an increased support from pro-vaccination groups \cite{10.1371/journal.pone.0247642}. Leading to the variance in cosine distance ($\sim$50\%), hence showcasing the growing distrust in the organization.

\subsection{Case Study : COVID-19 Trends}

In this instance, the model created in the previous example is further extended and built upon on, where now instead of equating a single relative term to the base term, multiple relative terms are considered. Therefore, not only analyzing their trends with respect to the base term but with themselves as well. Now, considering the base term as ``Trust'' while examining the association of this term with relative terms such as ``CDC'', ``Fauci'' and ``Experts''.

\begin{table}[ht]
    \centering
    \captionsetup{justification=centering}
    \begin{tabular}{|c|c|c|c|c|c|c|}
        \hline
        \multicolumn{7}{|c|}{Base Term : \textbf{Trust}}\\
        \hline
        Relative Term & \multicolumn{6}{c|}{Timeline (2020 - 2021)}\\
        \hline
        Months &  Jan'20 & Jun'20 & Dec'20 &  Jan'21 & Jun'21 & Dec'21\\
        \hline
        \hline
        CDC &  \cellcolor{blue!25}00.00 & \cellcolor{red!25}0.413 & 0.132 & 0.128 & \cellcolor{green!25}0.103 & 0.114\\
        \hline
        Fauci &  \cellcolor{blue!25}00.00 & \cellcolor{red!25}0.521 & 0.187 & \cellcolor{green!25}0.111 & 0.139 & 0.153\\
        \hline
        Experts &  0.280 & \cellcolor{red!25}0.591 & 0.133 & 0.126 & 0.146 & \cellcolor{green!25}0.109\\
        \hline
    \end{tabular}
    \caption{Cosine Similarity between COVID-19 topics over time in 2020-21.  The maximum Cosine Value is red whereas the minimum is green.}
    \label{tab:tab2}
\end{table}

\begin{figure}[H]
    \centering
    \captionsetup{justification=centering}
    \includegraphics[width=1\textwidth]{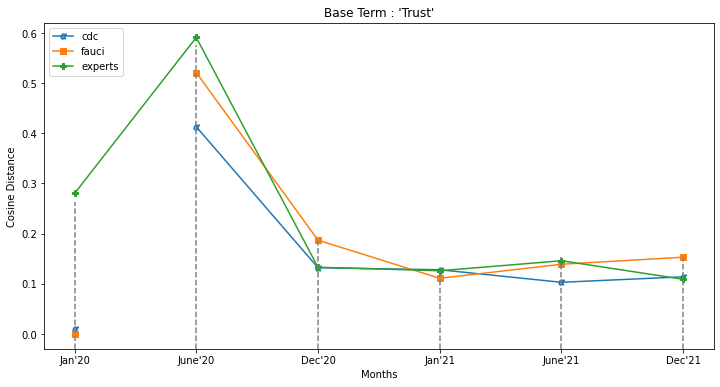}
    \caption{Trend of the terms ``CDC'', ``Fauci'' and ``experts'' through 2020-21 with the relative term ``Trust''}
    \label{fig:graph2}
\end{figure}

The following case study is a perfect example to signify the discrepancies observed by quantizing such arbitrary concepts such as trust and formalism of an organization. Even though the terms ``Unofficial'' (refer to Figure \ref{fig:graph1} and Table \ref{tab:tab1}) and ``Trust'' (refer to Figure \ref{fig:graph2} and Table \ref{tab:tab2}) may seem different semantically, both of them produce similar trends when in relation to ``CDC''. This may be attributed to the absence of a general consensus. 

An important fact that can be observed in both cases is the large difference in the values of the terms between June 2020 to December 2020. This corresponds to a greater shift in the political landscape as well as the public opinion. Where one school of thought went from considering CDC from competent to inept, in contrast, another may have started considering the said organization as more trustworthy. But, the COVID-19 case study also displays that all the relative terms (that are closely correlated to each-other) such as CDC, Dr. Anthony Fauci (Chief Medical Advisor to the President of the United States) and Experts all show similar trends.

\subsection{Case Study : The Olympics Trends}

While the previous example showed an exception to these trends, a notable demonstration of the inference power of such trends can be observed by taking the example of the Olympic games. As these games are periodic, occurring every 4 years, have very specific sub-topics (certain sports that are conducted at the Olympics) and do not have differentiating opinions about them, they are a perfect fit for the present study.

The base term is set as ``Olympics'' in the models and the relative terms were kept as pivotal sporting events, ``Track'', ``Tennis'', ``Gymnastics'' and ``Race'', as well as certain sports that are not a part of the Olympics (and hence not closely related) such as American Football as the relative term ``Football''. 

These trends were drawn over 3 years (2016, 2017, 2021) where only 2017 was a non-Olympic year. Out of each of these years, 5 months were considered (January, March, July, September and December) to draw out these trends from.

\subsubsection{2016 Trends}

\begin{table}[H]
    \centering
    \captionsetup{justification=centering}
    \begin{tabular}{|c|c|c|c|c|c|c|}
        \hline
        \multicolumn{6}{|c|}{Base Term : \textbf{Olympics}}\\
        \hline
        Relative Term & \multicolumn{5}{c|}{Timeline (2016)}\\
        \hline
        Months &  Jan'16 & Mar'16 & Jul'16 &  Sep'16 & Dec'16\\
        \hline
        \hline
        Track & \cellcolor{red!25}0.368 & 0.355 & \cellcolor{green!25}0.115 & 0.215 & 0.309 \\
        \hline
        Football & 0.632 & \cellcolor{red!25}0.658 & \cellcolor{green!25}0.589 & 0.633 & 0.641\\
        \hline
        Tennis & \cellcolor{red!25}0.563 & 0.418 & \cellcolor{green!25}0.234 & 0.380 & 0.518\\
        \hline
        Gymnastics &  0.433 & \cellcolor{red!25}0.456 & \cellcolor{green!25}0.218 & 0.373 & 0.439\\
        \hline
        Race &  \cellcolor{red!25}0.878 & 0.675 & \cellcolor{green!25}0.268 & 0.480 & 0.780\\
        \hline
    \end{tabular}
    \caption{Cosine Similarity between Sports topics over time in 2016. Maximum cosine value is red whereas the Minimum cosine value is green.}
    \label{tab:tab3}
\end{table}

\begin{figure}[H]
    \centering
    \captionsetup{justification=centering}
    \includegraphics[width=1\textwidth]{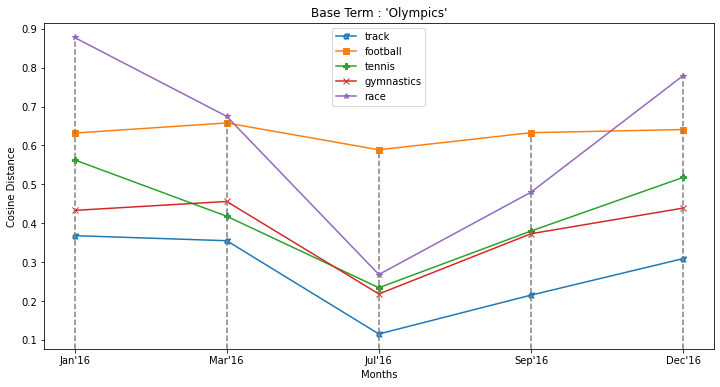}
    \caption{Trend of the terms ``track'', ``football'', ``tennis'', ``gymnastics'' and ``race'' through 2016 with the relative term ``Olympics''}
    \label{fig:graph3}
\end{figure}

Using the data present in sports articles and analyzing the trends of such sports as track, tennis, gymnastics (all relating to and conducted at the Olympics) similar trends are shown. Whereas, American Football (not related to the Olympics) unlike the rest of the relative terms remains a nearly constant and static trend.

Moreover, the 2016 Summer Olympics, (also known as Rio 2016) was held from \nth{5} to \nth{21} August 2016 in Rio de Janeiro, Brazil, with preliminary events in some sports beginning on \nth{3} August. This sharply coincides with the results, where all relative terms related to Olympics show the highest convergence in the month of July 2016.

\subsubsection{2017 Trends}

\begin{table}[H]
    \centering
    \captionsetup{justification=centering}
    \begin{tabular}{|c|c|c|c|c|c|c|}
        \hline
        \multicolumn{6}{|c|}{Base Term : \textbf{Olympics}}\\
        \hline
        Relative Term & \multicolumn{5}{c|}{Timeline (2017)}\\
        \hline
        Months &  Jan'17 & Mar'17 & Jul'17 &  Sep'17 & Dec'17\\
        \hline
        \hline
        Track &  0.344 & 0.361 & \cellcolor{green!25}0.337 & \cellcolor{red!25}0.388 & 0.362 \\
        \hline
        Football & \cellcolor{green!25}0.622 & 0.673 & \cellcolor{red!25}0.681 & 0.641 & 0.679\\
        \hline
        Tennis &  \cellcolor{red!25}0.587 & 0.568 & 0.543 & 0.562 & \cellcolor{green!25}0.418\\
        \hline
        Gymnastics &  0.454 & 0.456 & \cellcolor{red!25}0.531 & \cellcolor{green!25}0.357 & 0.459\\
        \hline
        Race &  \cellcolor{green!25}0.678 & \cellcolor{red!25}0.875 & 0.743 & 0.756 & 0.866\\
        \hline
    \end{tabular}
    \caption{Cosine Similarity between Sports topics over time in 2017. The Maximum cosine value is red whereas the Minimum cosine value is green.}
    \label{tab:tab4}
\end{table}

\begin{figure}[H]
    \centering
    \captionsetup{justification=centering}
    \includegraphics[width=1\textwidth]{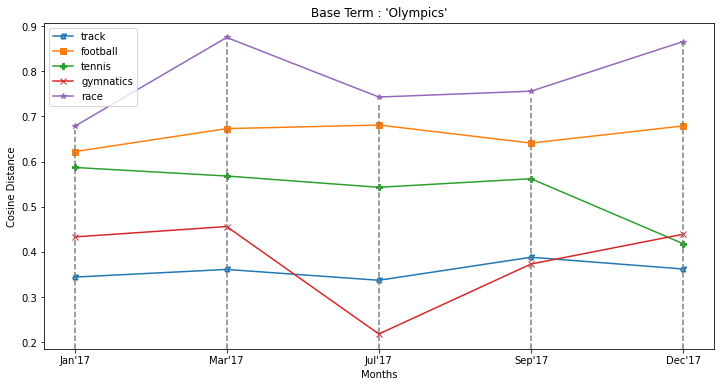}
    \caption{Trend of the terms ``track'', ``football'', ``tennis'', ``gymnastics'' and ``race'' through 2017 with the relative term ``Olympics''}
    \label{fig:graph4}
\end{figure}

Now, using the same data from sports articles and other miscellaneous sources in 2017, trends are generated using the same relative and base term. All trends (refer to Table \ref{tab:tab4}) still show variation over time but do not show the same convergence as observed in the sports trends for 2016 (refer to Table \ref{tab:tab5}). Instead, all trends follow stable and consistent paths, exhibiting minimal variation.

An important contextual distinction that can be discerned is the fact that 2017 is a year in which Olympics are not held in. The crest and trough of the trends may very well show some particular event that relates to the sport, hence affecting its relative term. However, as there are no specific entities common between these, no particular trends are observed. 

\subsubsection{2021 Trends}

\begin{table}[H]
    \centering
    \captionsetup{justification=centering}
    \begin{tabular}{|c|c|c|c|c|c|c|}
        \hline
        \multicolumn{6}{|c|}{Base Term : \textbf{Olympics}}\\
        \hline
        Relative Term & \multicolumn{5}{c|}{Timeline (2021)}\\
        \hline
        Months &  Jan'21 & Mar'21 & Jul'21 &  Sep'21 & Dec'21\\
        \hline
        \hline
        Track &  \cellcolor{red!25}0.353 & 0.328 & \cellcolor{green!25}0.109 & 0.263 & 0.312 \\
        \hline
        Football & 0.656 & \cellcolor{red!25}0.681 & \cellcolor{green!25}0.597 & 0.624 & 0.638\\
        \hline
        Tennis &  \cellcolor{red!25}0.573 & 0.410 & 0.244 & \cellcolor{green!25}0.210 & 0.318\\
        \hline
        Gymnastics &  0.212 & 0.349 & \cellcolor{green!25}0.218 & 0.398 & \cellcolor{red!25}0.441\\
        \hline
        Race &  \cellcolor{red!25}0.900 & 0.576 & \cellcolor{green!25}0.254 & 0.503 & 0.823\\
        \hline
    \end{tabular}
    \caption{Cosine Similarity between Sports topics over time in 2021. The Maximum cosine value is red whereas the Minimum cosine value is green.}
    \label{tab:tab5}
\end{table}

\begin{figure}[H]
    \centering
    \captionsetup{justification=centering}
    \includegraphics[width=1\textwidth]{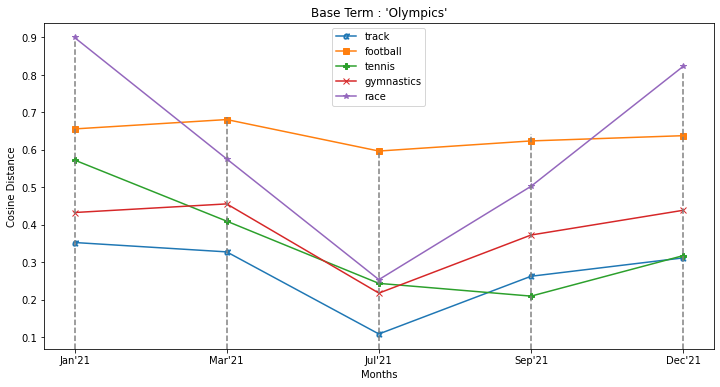}
    \caption{Trend of the terms ``track'', ``football'', ``tennis'', ``gymnastics'' and ``race'' through 2021 with the relative term ``Olympics''}
    \label{fig:graph5}
\end{figure}

Finally, we see a great deal of similarity between the trends of 2021 (refer to Table \ref{tab:tab5}) and 2016 (refer to Table \ref{tab:tab3}). This shows that the trends generated in the present study are reproducible and show periodicity, where the trends of 2017 are visibly differentiable from 2016 and 2021.

The 2020 Summer Olympics, (also known as Tokyo 2020), was held from \nth{23} July to \nth{8} August 2021 in Tokyo, Japan, with some preliminary events that began on \nth{21} July, 2021. Similar to 2016, this sharply coincides with the results, where all relative terms related to Olympics show the highest convergence in the month of July 2016.

\newpage

\section{Results and Discussion}
The present study has discussed the limitations of using stochastic methods to try to understand how language and the topics within it are transformed over time. Thus, introducing a method of using Word2Vec to capture semantic meaning to draw out changes in the meaning and use of words over time, and therefore understanding and visualizing the trends related to them, has been helpful. We have presented two case studies to substantiate our hypothesis.

The model shows how articles pertaining to sports show multiple trends of certain terms with respect to sports itself. But, it is none the more apparent with competitive games that are and are not related to the Olympics. Sports like track \& field, racing, tennis, gymnastics show a trend of higher correlation with sports during the months around June, compared to American Football, (refer to Figure \ref{fig:graph3}) which is not an Olympic sport, to show no such trend and remaining mostly non-divergent. Whereas track \& field, arguably the sport most dependent on the Olympics, shows the highest deviation compared to other sports such as tennis, which has other important events all year around. 

Another case study relates to terms associated with healthcare during the COVID-19 pandemic, where we consider the trend of terms with respect to the dimension of ``Trust'' and ``Unofficial''. The first cases of COVID were reported in China in the first weeks of January 2020. This shows the center of disease control, ``CDC'' with a comparatively low correlation to trust (refer to Figure \ref{fig:graph2}) as well as unofficial (refer to Figure \ref{fig:graph1}). Germani et al \cite{10.1371/journal.pone.0247642}, in their behavioural analysis research of anti-vaccination rhetoric on social media, depict a lot of similarity with the trends in the present study.

\section{Conclusion \& Future Work}

The present study presents a new method to measure the temporal trend of words by capturing the change in their context. The semantic movement of a target topical word is tracked and compared against a base term with the cosine similarity of the word vectors taken over time, measuring how topical keywords can evolve. This approach goes beyond the current explored methods which only track single keyword occurrences over time. Hence, this approach also provides a better means to visualize the trends of topics over time.

As a standalone tool, this framework can assist in understanding the evolutionary aspects language undergoes as words can gain and lose meaning over time. This is especially true in the social networks where the primary mode of communication is text, and discourse occurs quite frequently and rapidly. In the presented two case studies, the domains of healthcare and sports with a news dataset were explored, both exhibiting visible discernment.

Combined with automated topic extraction methods, such as LDA \cite{10.5555/944919.944937} and Top2Vec \cite{angelov2020top2vec} , topic discovery could be enhanced with topical evolution as an end-to-end framework for developing deeper insights into a corpus. This includes inter-topic trends as words can be part of a topic in one time period, and then be a part of another.

The current approach utilizes multiple models, each corresponding to a specified time period. Training a corpus on a single model preserves the basis space of the embeddings as to minimize the correlation errors, and in our case, between paired topics. The hope is to develop a single model that jointly trains the embeddings with respect to the time aspects of the corpus could solve this issue. However, as the intention was to develop a practical analytical tool since time-based learning is limited in current language models, careful construction of the dataset is a suitable solution. Whether it be by joining all the month datasets together and tagging each word with its month, or by creating intermediate datasets between each month by virtue of an an overlapping window.

Recently, more robust deep learning architectures that yield improved embeddings due to bidirectional learning (e.g. BERT \cite{devlin2018bert} or ELMo \cite{sarzynska2021detecting}). As these language models can learn different usages and interpretations of a token, our framework would be less prone to errors in the presence of linguistic polysemy.

\newpage

\printbibliography

\end{document}